%% file: MV.tex
\documentclass{article}
\input{MinhPreamble.tex}

\title{Gaussian-Mixture-Model Q-functions for Reinforcement
  Learning\\ by Riemannian Optimization \vspace{-30pt}
}

\name{}

\address{
  \begin{minipage}{.7\textwidth}
    \begin{center}
      \textit{Minh Vu\qquad Konstantinos Slavakis} \\[1ex]
      \small Tokyo Institute of Technology, Japan\\
      Department of Information and Communications Engineering\\
      Emails: \texttt{\{vu.d.aa, slavakis.k.aa\}@m.titech.ac.jp}
    \end{center}
  \end{minipage}
}

\begin{document}
\ninept

\maketitle

\begin{abstract}
  This paper establishes a novel role for Gaussian-mixture
  models (GMMs) as functional approximators of Q-function
  losses in reinforcement learning (RL). Unlike the existing
  RL literature, where GMMs play their typical role as
  estimates of probability density functions, GMMs
  approximate here Q-function losses. The new Q-function
  approximators, coined GMM-QFs, are incorporated in Bellman
  residuals to promote a Riemannian-optimization task as a
  novel policy-evaluation step in standard policy-iteration
  schemes. The paper demonstrates how the hyperparameters
  (means and covariance matrices) of the Gaussian kernels
  are learned from the data, opening thus the door of RL to
  the powerful toolbox of Riemannian optimization. Numerical
  tests show that with no use of experienced data, the proposed
  design outperforms state-of-the-art methods, even deep
  Q-networks which use experienced data, on benchmark RL tasks.
\end{abstract}


\section{Introduction}\label{sec:intro}

In reinforcement learning (RL)~\cite{Bertsekas:RLandOC:19,
  Sutton:IntroRL:18}, an ``intelligent agent'' interacts
with an unknown environment (typically modeled as a Markov
decision process (MDP)) to identify an optimal policy that
minimizes the total costs of its ``actions.''  RL offers a
mathematically sound framework for solving arduous
sequential decision problems in real-world applications, as
in operations research, dynamic control, data mining, and
bioinformatics~\cite{Bertsekas:RLandOC:19}.

To identify an optimal policy, RL strategies typically
compute/evaluate the \textit{value/loss}\/ (Q-function)
associated with an action at a given state by observing
feedback data from the environment. The classical
Q-learning~\cite{watkins92Qlearning} and
state-action-reward-state-action (SARSA)~\cite{singh00sarsa}
algorithms evaluate Q-functions by look-up tables, populated
by Q-function values at \textit{every possible}\/
state-action pair. Although such approaches appear to be
successful in discrete-state-action RL, there are many
practical problems which involve very large, or even
continuous state-action spaces that render tabular RL
methods computationally intractable. To overcome this
difficulty, algorithms built on functional approximations
(non-linear models) of Q-functions have attracted
considerable interest~\cite{Bertsekas:RLandOC:19}.



Functional approximations of Q-functions have a long
history in RL. Classical kernel-based (KB)RL
methods~\cite{ormoneit02kernel, ormoneit:autom:02,
  bae:mlsp:11} model Q-functions as elements of Banach
spaces; usually, spaces comprising all essentially bounded
functions. On the other hand, temporal difference
(TD)~\cite{sutton88td}, least-squares
(LS)TD~\cite{lagoudakis03lspi, regularizedpi:16, xu07klspi},
Bellman-residual (BR) methods~\cite{onlineBRloss:16}, as
well as very recent nonparametric designs~\cite{vu23rl,
  akiyama24proximal, akiyama24nonparametric}, model
Q-functions as elements of user-defined reproducing kernel
Hilbert spaces (RKHSs)~\cite{aronszajn50kernels,
  scholkopf2002learning} in a quest to exploit the geometry
and computational convenience of the associated inner
product and its reproducing property. Notwithstanding, the
number of design parameters of all of the aforementioned
kernel-based designs scale with the number of observed data,
which usually inflicts memory and computational bottlenecks
when operating in dynamic environments with time-varying
data distributions. Dimensionality reduction techniques have
been introduced to address this issue~\cite{xu07klspi,
  vu23rl}, but reducing the number of basis elements of the
approximating subspace may hinder the quality of the
Q-functions estimates.


Deep neural networks have been also used as non-linear
Q-function approximators in the form of deep Q-networks
(DQNs), e.g.,~\cite{mnih13dqn, hasselt16ddqn}. Typically,
DQN models require experienced data~\cite{lin93experience} from past
policies for their parameters
to be learned, and may even require re-training during online mode
to learn from data with probability density functions (PDFs)
which are different from those of the past data
(experience-replay buffer). Such requirements may yield
large computational times and complexity footprints,
discouraging the application of DQNs into online learning
where lightweight operations and swift adaptability to a
dynamic environment are desired.


Aiming at a novel class of Q-function estimates with rich
approximating properties, with few parameters to be learned
to effect dimensionality reduction, robustness to erroneous
information, and swift adaptability to dynamic environments,
and with \textit{no}\/ need for past experienced data, this paper
introduces the class of \textit{Gaussian-mixture-model
  Q-functions (GMM-QFs).} GMM-QFs are weighted sum averages
of multivariate Gaussian kernels, where not only the
weights, but also the hyperparameters of the Gaussian
kernels are free to be
learned~\cite{McLachlan:FiniteMixtures:00}. This contrasts
the aforementioned literature of
KBRL~\cite{ormoneit02kernel, ormoneit:autom:02, bae:mlsp:11,
  sutton88td, lagoudakis03lspi, regularizedpi:16, xu07klspi,
  onlineBRloss:16}, where the hyperparameters of the
user-defined kernels are directly parameterized by the
observed data and are not considered variables of learning
tasks. GMMs have been already used in RL, but via their
typical role as estimates of PDFs: either of the joint PDF
$p(Q, \vect{s}, a)$~\cite{sato99em, mannor05rl,
  agostini10gmmrl, agostini17gmmrl}, where the Q-function
$Q$, as well as state $\vect{s}$ and action $a$ are
considered to be random variables (RVs), or of the
conditional PDF
$p(Q \given \vect{s}, a)$~\cite{choi19distRL}. This
classical usage of GMMs and its intimate connection with
maximum-likelihood estimation~\cite{Demp:77,
  Figueiredo:mixtures:02} lead naturally to
expectation-maximization (EM) solutions~\cite{sato99em,
  mannor05rl, agostini10gmmrl, agostini17gmmrl}. In
contrast, this paper departs from the typical GMM usage and
their EM solutions~\cite{sato99em, mannor05rl,
  agostini10gmmrl, agostini17gmmrl, choi19distRL}, employs
GMMs to model Q-functions \textit{directly,} and not their
PDFs, follows the lines of Bellman-residual (BR)
minimization~\cite{onlineBRloss:16, akiyama24nonparametric}
to form a smooth objective function, and relies on
Riemannian optimization~\cite{Absil:OptimManifolds:08} to
minimize that objective function and to exploit the
underlying Riemannian geometry~\cite{RobbinSalamon:22} of
the hyperparameter space. GMMs and Riemannian optimization
have been used to model policy functions as
PDFs~\cite{RPPO}, under the framework of policy
search~\cite{sutton99pg}. The use of GMMs to model
Q-functions directly via Riemannian optimization seems to
appear here for the \textit{first time}\/ in the RL
literature.

A fixed number of Gaussian kernels are used in GMM-QFs to
address the problem of an overgrowing nonparametric model
with the number of data~\cite{ormoneit02kernel,
  ormoneit:autom:02, bae:mlsp:11, sutton88td,
  lagoudakis03lspi, regularizedpi:16, xu07klspi,
  onlineBRloss:16}, effecting dimensionality reduction, and
providing low-computational load as well as stable
performance under erroneous information. Indeed, numerical
tests on benchmark control tasks demonstrate that the
advocated GMM-QFs outperform other state-of-the-art RL
schemes, even DQNs which require experienced data. Due to
limited space, detailed definitions and arguments of
Riemannian geometry, proofs, results on convergence, and
further numerical tests are deferred to the journal version
of this manuscript.

\section{The Class of GMM Q-Functions
  (GMM-QF\lowercase{s})}\label{sec:GMM-QFs}

\subsection{RL notations}\label{subsec:notations}

Let $\mathfrak{S} \subset \Real^D$ denote the
\textit{continuous} state space, with state vector
$\vect{s} \in \mathfrak{S}$, for some $D \in \IntegerPP$
($\IntegerPP$ is the set of all positive integers). The
usually discrete action space is denoted by $\mathfrak{A}$,
with action $a\in\mathfrak{A}$. An agent, currently at state
$\vect{s} \in \mathfrak{S}$, takes an action
$a\in \mathfrak{A}$ and transits to a new state
$\vect{s}^\prime \in\mathfrak{S}$ under transition
probability $p(\vect{s}^\prime \given \vect{s}, a)$ with an
\textit{one-step} loss $g(\vect{s},a)$. The Q-function
$Q(\cdot, \cdot) \colon \mathfrak{S} \times \mathfrak{A} \to
\Real \colon (\vect{s}, a) \mapsto Q(\vect{s}, a)$ stands
for the \textit{long-term}\/ loss/cost that the agent will
suffer/pay, if the agent takes action $a$ at state
$\vect{s}$. For convenience, the state-action tuple
$\vect{z} \coloneqq (\vect{s},a) \in \mathfrak{Z} \coloneqq
\mathfrak{S} \times \mathfrak{A} \subset \Real^{D_z}$, where
$D_z \in \IntegerPP$.

Following~\cite{Bertsekas:RLandOC:19}, consider the set of
all mappings
$\mathcal{M} \coloneqq \Set{\mu(\cdot) \given \mu(\cdot)
  \colon \mathfrak{S} \to \mathfrak{A} \colon \vect{s}
  \mapsto \mu(\vect{s})}$. In other words, $\mu(\vect{s})$
denotes the action that the agent will take at state
$\vect{s}$ under $\mu$. The set of policies is defined as
$\Pi \coloneqq \mathcal{M}^{\IntegerP} \coloneqq \Set{\mu_0,
  \mu_1, \dots, \mu_n, \dots \given \mu_n \in \mathcal{M}, n
  \in \IntegerP}$. A policy will be denoted by
$\pi \in \Pi$. Given $\mu\in\mathcal{M}$, a stationary
policy $\pi_\mu$ is defined as
$\pi_\mu \coloneqq (\mu, \mu, \dots, \mu, \dots)$. It is
customary for $\mu$ to denote also $\pi_\mu$.

\subsection{GMM-QFs}

Motivated by GMMs~\cite{McLachlan:FiniteMixtures:00}, and
for a user-defined positive integer $K$, GMM-QFs are defined
as the following class of functions:
\begin{align}
  \mathcal{Q} \coloneqq \Bigl\{ Q(\vect{z})
  & \coloneqq
    \sum\nolimits_{k=1}^K \xi_k \mathscr{G}(\vect{z} \mid
    \vect{m}_k, \vect{C}_k) \mathop{} \big|
    \mathop{} \xi_k \in \Real\,,
    \vect{m}_k \in \Real^{D_z}\,, \notag \\
  & \Real^{D_z \times D_z} \ni\vect{C}_k\ \text{is positive
    definite}\,, \forall k = 1, \ldots, K \Bigr\}
    \,, \label{eq:gmm-q}
\end{align}
where
$\mathscr{G}(\vect{z} \mid \vect{m}_k, \vect{C}_k) \coloneqq
\exp[ - (\vect{z} - \vect{m}_k)^{\intercal} \vect{C}_k^{-1}
(\vect{z} - \vect{m}_k)]$, with $\vect{m}_k$ and
$\vect{C}_k$ being the hyperparameters of
$\mathscr{G}(\cdot)$, widely known as the ``mean'' and
``covariance matrix'' of $\mathscr{G}(\cdot)$,
respectively, while $\intercal$ stands for vector/matrix
transposition. The parameter space of GMM-QFs takes the form
\begin{alignat}{3}
  \mathscr{M}
  & {} \coloneqq {}
  && \Bigl\{ \vectgr{\Omega}
     && \coloneqq (
     \xi_1, \ldots, \xi_K, \vect{m}_1, \ldots,
     \vect{m}_K, \vect{C}_1, \ldots,
        \vect{C}_K)  \mathop{} \big|
    \mathop{} \xi_k \in \Real\,, \notag\\
  &&&&& \vect{m}_k \in \Real^{D_z}\,,
     \vect{C}_k\ \text{is positive definite}\,,
     \forall k = 1, \ldots, K
     \Bigr\} \notag \\
  & = && \Real^K && \times \Real^{D_z \times K} \times
         ( \mathbb{S}_{++}^{D_z} )^K \,, \label{param.space}
\end{alignat}
where $\mathbb{S}_{++}^{D_z}$ stands for the set of all
$D_z\times D_z$ positive-definite matrices. Interestingly,
$\mathscr{M}$ is a Riemannian
manifold~\cite{RobbinSalamon:22, Absil:OptimManifolds:08}
because all of $\Real^K$, $\Real^{D_z \times K}$, and
$\mathbb{S}_{++}^{D_z}$ are.

To learn the ``optimal'' parameters
from~\eqref{param.space}, BR
minimization~\cite{onlineBRloss:16, qin14sparseRL,
  mahadevan14proximal, liu18proximalGTD,
  akiyama24nonparametric} is employed. Motivation comes
from the classical Bellman
mappings~\cite{Bertsekas:RLandOC:19}, which quantify the
total loss ($=$ one-step loss $+$ expected long-term loss)
to be paid by the agent, had action $a$ been taken at state
$\vect{s}$. More specifically, if $\Banach$ stands for the
space of Q-functions, usually being the Banach space of all
essentially bounded functions~\cite{Bertsekas:RLandOC:19},
then the classical Bellman mappings
$T^{\diamond}_\mu, T^{\diamond} \colon \Banach \to \Banach
\colon Q \mapsto T^{\diamond}_\mu Q, T^{\diamond} Q$ are
defined as~\cite{Bertsekas:RLandOC:19}
\begin{subequations}\label{Bellman.maps.standard}
  \begin{align}
    (T_{\mu}^{\diamond} Q)(\vect{s}, a)
    & \coloneqq g( \vect{s}, a ) + \alpha
      \mathbb{E}_{\vect{s}^{\prime} \given
      (\vect{s}, a)} [ Q(\vect{s}^{\prime},
      \mu(\vect{s}^{\prime}))
      ]\,, \label{Bellman.standard.mu} \\
    (T^{\diamond} Q)(\vect{s}, a)
    & \coloneqq g( \vect{s}, a ) + \alpha
      \mathbb{E}_{\vect{s}^{\prime} \given
      (\vect{s}, a)} [ \min\nolimits_{a^{\prime}\in
      \mathfrak{A}} Q(\vect{s}^{\prime}, a^{\prime})
      ]\,, \label{Bellman.standard}
  \end{align}
\end{subequations}
$\forall (\vect{s}, a)$, where
$\mathbb{E}_{\vect{s}^{\prime}\given (\vect{s},a)}[\cdot]$
stands for the conditional expectation operator with respect
to the potentially next state $\vect{s}^{\prime}$
conditioned on $(\vect{s},a)$, and $\alpha \in [0,1)$ is the
discount factor. Mapping \eqref{Bellman.standard.mu} refers
to the case where the agent takes actions according to the
stationary policy $\mu$, while \eqref{Bellman.standard}
serves as a greedy variation of \eqref{Bellman.standard.mu}.

Given mapping $T: \Banach \to \Banach$, its fixed-point set
$\Fix T\coloneqq \{Q\in\Banach \given TQ=Q \}$. It is
well-known that the fixed-point sets $\Fix T_{\mu}^\diamond$
and $\Fix T^\diamond$ play central roles in identifying
\textit{optimal}\/ policies which minimizes the total
loss~\cite{Bertsekas:RLandOC:19}. Usually, the discount
factor $\alpha \in [0,1)$
to render
$T_{\mu}^\diamond,~T^\diamond$ strict
contractions~\cite{Bertsekas:RLandOC:19, hb.plc.book};
hence, $\Fix T_{\mu}^\diamond$ and $\Fix T^\diamond$ become
singletons. It is clear from~\eqref{Bellman.maps.standard}
that the computation of $\Fix T_{\mu}^\diamond$ and
$\Fix T^\diamond$ requires the knowledge on the transition
probabilities to be able to compute the conditional
expectation
$\mathbb{E}_{\vect{s}^{\prime} \given (\vect{s}, a)}[ \cdot
]$.  However, in most cases of practice, transition
probabilities are unavailable to the agent. To surmount this
lack of information, designers utilize models for
Q-functions. This manuscript utilizes GMM-QFs
in~\eqref{eq:gmm-q}.

Motivated by the importance of fixed points of Bellman
mappings in RL, and for the data samples
$\mathcal{D}_{\mu} \coloneqq \Set{ (\vect{s}_t, a_t, g_t,
  \vect{s}_t^\prime) }_{t=1}^T$, for a number $T$ of time
instances under a stationary policy $\mu$, the following
minimization task of the smooth objective
$\mathscr{L}(\cdot)$ over the Riemannian manifold
$\mathscr{M}$ will be used to identify the desired
fixed-point Q-functions for the policy $\mu$:
\begin{align}
  \min_{\vectgr{\Omega} \in \mathscr{M}}
  \mathscr{L}(\vectgr{\Omega}) \coloneqq
  \sum\nolimits_{t=1}^{T} \Bigl[ g_t
  & + \alpha
  \sum\nolimits_{k=1}^{K}\xi_k
  \mathscr{G}(\vect{z}^\prime_t \given
    \vect{m}_k,\vect{C}_k) \notag\\
  & - \sum\nolimits_{k=1}^{K}\xi_k
  \mathscr{G}(\vect{z}_t \given \vect{m}_k,\vect{C}_k)
    \Bigr]^2 \,, \label{parameters.task}
\end{align}
where $\vect{z}_t \coloneqq (\vect{s}_t, a_t)$ and
$\vect{z}_t^{\prime} \coloneqq ( \vect{s}_t^{\prime},
\mu(\vect{s}_t^{\prime}) )$. Task~\eqref{parameters.task} is
solved by \cref{algo:armijo}.

Albeit the similarity of~\eqref{parameters.task} with
standard BR minimization~\cite{onlineBRloss:16,
  qin14sparseRL, mahadevan14proximal, liu18proximalGTD,
  akiyama24nonparametric}, \eqref{parameters.task} is
performed over a parameter space, parameterized not only by
the weights $\vectgr{\xi}$, as in~\cite{onlineBRloss:16,
  qin14sparseRL, mahadevan14proximal, liu18proximalGTD,
  akiyama24nonparametric}, but also by the parameters
$\{ \vect{m}_k, \vect{C}_k \}_{k=1}^K$. In other words, and
for a fixed $K$, \eqref{parameters.task} provides more
degrees of freedom and a richer parameter space than the
state-of-the-art BR-minimization
methods~\cite{onlineBRloss:16, qin14sparseRL,
  mahadevan14proximal, liu18proximalGTD,
  akiyama24nonparametric}.

\section{Policy iteration by\\ Riemannian
  optimization}\label{sec:rpi}

Following standard routes~\cite{Bertsekas:RLandOC:19,
  konda99ac}, the classical policy-iteration (PI) strategy
is used in \cref{algo:PI} to identify optimal policies. PI
comprises two stages per iteration $n$: \textit{policy
  evaluation} and \textit{policy improvement.} At policy
evaluation, the current policy is evaluated by the current
Q-function estimate, which represents the long-term
cost/loss estimate that the agent would suffer had the
current policy been used to determine the next state. At the
policy-improvement stage, the agent uses the obtained
Q-function values to update the policy.

Nevertheless, looking more closely at
\cref{algo:update.omega} of \cref{algo:PI}, the
policy-evaluation stage is \textit{newly}\/ equipped here
with a Riemannian-optimization task:
solve~\eqref{parameters.task} by the
steepest-gradient-descent method with line search
of~\cite[\S4.6.3]{Absil:OptimManifolds:08}. To this end, the
gradients of $\mathscr{L}( \cdot )$ along the directions
$\vectgr{\xi}$, $\vect{m}_k$ and $\vect{C}_k$ are required,
and provided by \cref{prop:gradients}. Definitions of basic
Riemannian concepts~\cite{Absil:OptimManifolds:08,
  RobbinSalamon:22}, detailed derivations and proofs are
skipped because of limited space.

\begin{algorithm}[t!]
  \begin{algorithmic}[1]
    \renewcommand{\algorithmicindent}{1em}

    \State{Arbitrarily initialize $\vectgr{\Omega}_0 \in
      \mathscr{M}$, $\mu_0 \in \mathcal{M}$.}

    \While{$n \in \IntegerP$} \label{line:iter}

      \State{\textbf{Policy evaluation} Use the current
        policy $\mu_n$ to generate the dataset
        $\mathcal{D}_{\mu_n} \coloneqq \Set{(\vect{s}_t, a_t,
          g_t,\vect{s}_t^\prime)}_{t=1}^{T}$.}

      \State{Update $\vectgr{\Omega}_{n+1}$
        via~\cref{algo:armijo}.}\label{algo:update.omega}

      \State{Given $\vectgr{\Omega}_{n+1}$, compute $Q_{n+1}
        \in \mathcal{Q}$
        via~\eqref{eq:gmm-q}.}\label{algo:ROforPE}

      \State{\textbf{Policy improvement} Update $\mu_{n+1}
        \coloneqq \underaccent{a\in \mathfrak{A}}{\arg\min}
        Q_{n+1}(\vect{s},a)$.
        }

      \State{Increase $n$ by one, go to~\cref{line:iter}.}

    \EndWhile
  \end{algorithmic}
  \caption{Policy iteration by Riemannian
    optimization} \label{algo:PI}
\end{algorithm}

To run computations in \cref{algo:armijo}, the Riemannian
metric~\cite{Absil:OptimManifolds:08, RobbinSalamon:22}
of~\eqref{Rmetric.on.M} on $\mathscr{M}$ is adopted:
$\forall \vectgr{\Omega} \coloneqq ( \vectgr{\xi},
\vect{m}_{1}, \dots, \vect{m}_{K}, \vect{C}_{1}, \dots,
\vect{C}_{K} ) \in \mathscr{M}$, and
$\forall \vectgr{\Upsilon}_i \coloneqq ( \vectgr{\theta}_i,
\vectgr{\mu}_{i1}, \dots, \vectgr{\mu}_{iK},
\vectgr{\Gamma}_{i1}, \dots, \vectgr{\Gamma}_{iK} ) \in
T_{\vectgr{\Omega}} \mathscr{M}$, $i=1, 2$, where
$T_{\vectgr{\Omega}} \mathscr{M}$ denotes the tangent space
to $\mathscr{M}$ at
$\vectgr{\Omega}$~\cite{Absil:OptimManifolds:08,
  RobbinSalamon:22},
\begin{align}
  \innerp{ \vectgr{\Upsilon}_1}
  { \vectgr{\Upsilon}_2 }_{
  \vectgr{\Omega} } \coloneqq
  \vectgr{\theta}_1^{\intercal}
  \vectgr{\theta}_2 +
  \sum_{k=1}^K \vectgr{\mu}_{1k}^{\intercal}
  \vectgr{\mu}_{2k} + \sum_{k=1}^K
  \innerp{ \vectgr{\Gamma}_{1k} }{ \vectgr{\Gamma}_{2k}
  }_{ \vect{C}_k } \,, \label{Rmetric.on.M}
\end{align}
where $\innerp{ \cdot }{ \cdot }_{ \vect{C}_k }$ can be any
user-defined Riemannian metric of $\PS^{D_z}$. Here, the
Bures-Wasserstein (BW) metric~\cite{bhatia19bw}
of~\eqref{eq:bw.metric} is used, because of its excellent
performance in numerical tests:
$\forall \vect{C}_k \in \PS^{D_z}$, and
$\forall \vectgr{\Gamma}_{ik} \in T_{\vect{C}_k}
\PS^{D_z}$, $i=1, 2$,
\begin{align}
  \innerp{ \vectgr{\Gamma}_{1k} }{
  \vectgr{\Gamma}_{2k} }_{\vect{C}_k} \coloneqq \innerp{
  \vectgr{\Gamma}_{1k} }{ \vectgr{\Gamma}_{2k}
  }^{\textnormal{BW}}_{\vect{C}_k}
  \coloneqq \tfrac{1}{2} \trace
  [ L_{\vect{C}_k}(\vectgr{\Gamma}_{1k} )
  \vectgr{\Gamma}_{2k} ] \,, \label{eq:bw.metric}
\end{align}
where the Lyapunov operator $L_{\vect{C}_k}(\cdot)$
satisfies
$\vect{C}_k L_{\vect{C}_k}( \vectgr{\Gamma}_{ik} ) +
L_{\vect{C}_k}( \vectgr{\Gamma}_{ik} ) \vect{C}_k =
\vectgr{\Gamma}_{ik}$~\cite{bobiti16lyapunov}.
Other Riemannian metrics on $\PS^{D_z}$, such as the
affine-invariant~\cite{Pennec:Riemannian:19} or
Log-Cholesky~\cite{Bhatia:PD:07} ones can be used
in~\eqref{Rmetric.on.M}. Due to limited space, results
obtained after employing those metrics will be reported
elsewhere.

\begin{subequations}\label{all.gradients}
  \begin{proposition}[Computing
    gradients]\label{prop:gradients} Consider a point
    $\vectgr{\Omega}^{(j)} \coloneqq ( \vectgr{\xi}^{(j)},
    \vect{m}_1^{(j)}, \dots, \vect{m}_K^{(j)},
    \vect{C}_1^{(j)}, \dots, \vect{C}_K^{(j)}) \in
    \mathscr{M}$ (see \cref{algo:armijo}), and its associated
    GMM-QF $Q^{(j)}$. Let also
    $\delta_t \coloneqq g_t + \alpha Q^{(j)}
    (\vect{z}_t^\prime) - Q^{(j)} (\vect{z}_t)$. Then, the
    following hold true.

    \begin{thmlist}

    \item If the objective function in~\eqref{parameters.task}
      is recast as
      $\mathscr{L}( \vectgr{\Omega}^{(j)} ) = \norm{\vect{g} +
        \vectgr{\Delta} \vectgr{\xi}^{(j)} }^2$, where
      $\vect{g} \coloneqq [g_1, \dots, g_T]^\intercal$ and
      $\vectgr{\Delta}$ is a $T \times K$ matrix with entries
      $\vectgr{\Delta}_{tk} \coloneqq \alpha \mathscr{G}
      (\vect{z}_t^\prime \given \vect{m}_k^{(j)},
      \vect{C}_k^{(j)} ) - \mathscr{G} (\vect{z}_t \given
      \vect{m}_k^{(j)}, \vect{C}_k^{(j)})$, then,
      \begin{equation}\label{eq:dL.dxi}
        \frac{\partial\mathscr{L}}{\partial \vectgr{\xi}} (
        \vectgr{\Omega}^{(j)} ) = 2 \vectgr{\Delta}^\intercal
        (\vect{g} + \vectgr{\Delta} \vectgr{\xi}^{(j)} ) \,.
      \end{equation}

    \item $\forall k=1, \dots, K$,
      \begin{alignat}{2}
        & \frac{\partial \mathscr{L}}{\partial \vect{m}_k} (
          \vectgr{\Omega}^{(j)} ) && \notag\\
        & = \sum_{t=1}^{T} 4 \delta_t
          \xi_k^{(j)} ( \vect{C}_k^{(j)} )^{-1}
          \big[ && \alpha (\vect{z}^\prime_t -
                   \vect{m}_k^{(j)})
                   \mathscr{G}(\vect{z}_t^\prime \given
                   \vect{m}_k^{(j)}, \vect{C}_k^{(j)} ) \notag \\
        &&& - (\vect{z}_t - \vect{m}_k^{(j)})
            \mathscr{G} (\vect{z}_t \given \vect{m}_k^{(j)},
            \vect{C}_k^{(j)}) \big] \,. \label{eq:dL.dm}
      \end{alignat}

    \item Under the BW metric~\cite{bhatia19bw}, $\forall k=1,
      \dots, K$,
      \begin{align}
        \frac{\partial\mathscr{L} }{\partial \vect{C}_k} (
        \vectgr{\Omega}^{(j)} )
        & = \sum_{t=1}^{T}
          4 \delta_t \xi_k^{(j)} [ (\vect{C}_k^{(j)})^{-1}
          \vect{B}_{tk} + \vect{B}_{tk} (\vect{C}_k^{(j)})^{-1}
          ] \notag \\
        & \in T_{ \vect{C}_k^{(j)} } \PS^{D_z}
          \,, \label{eq:dL.dC}
      \end{align}
      where
      $\vect{B}_{tk} \coloneqq \alpha (\vect{z}^\prime_t -
      \vect{m}_k^{(j)} ) (\vect{z}^\prime_t - \vect{m}_k^{(j)}
      )^\intercal \mathscr{G} (\vect{z}^\prime_t \given
      \vect{m}_k^{(j)}, \vect{C}_k^{(j)} ) - (\vect{z}_t
      - \vect{m}_k^{(j)} ) (\vect{z}_t - \vect{m}_k^{(j)}
      )^\intercal \mathscr{G} (\vect{z}_t \given
      \vect{m}_k^{(j)}, \vect{C}_k^{(j)} )$.

    \end{thmlist}
  \end{proposition}
\end{subequations}

\begin{algorithm}[t!]
  \begin{algorithmic}[1]
    \renewcommand{\algorithmicindent}{1em}

    \State{%
      \textbf{Require:} Sampled data $\mathcal{D}_{\mu_n}
      \coloneqq \Set{(\vect{s}_t, a_t,
        g_t,\vect{s}_t^\prime)}_{t=1}^{T}$; scalars
      $\bar{\alpha}>0, \beta\in (0,1), \sigma \in (0,1)$,
      the number of steps $J$, a
      Riemannian metric $\innerp{\cdot}{\cdot}_{\cdot}$, and
      a retraction mapping $R_{\cdot}(\cdot)$ on
      $\mathscr{M}$.
    }

    \State{%
      $\vectgr{\Omega}^{(0)} \coloneqq \vectgr{\Omega}_n$.
    }

    \For{$j = 0, 1, 2, \dots, J-1$}

      \State{%
        $\vectgr{\Omega}^{(j)}
        \coloneqq ( \vectgr{\xi}^{(j)}, \vect{m}_1^{(j)},
        \dots, \vect{m}_K^{(j)}, \vect{C}_1^{(j)}, \dots,
        \vect{C}_K^{(j)})$.
      }

      \State{%
        By~\eqref{all.gradients}, compute:
        \[
          \resizebox{.95\columnwidth}{!}{%
            $\nabla \mathscr{L} ( \vectgr{\Omega}^{(j)} ) = (
            \frac{ \partial\mathscr{L} } { \partial
              \vectgr{\xi} } ( \vectgr{\Omega}^{(j)} ), \dots,
            \frac{ \partial\mathscr{L} } { \partial \vect{m}_k
            } ( \vectgr{\Omega}^{(j)} ),
            \dots, \frac{ \partial\mathscr{L} } { \partial
              \vect{C}_k } ( \vectgr{\Omega}^{(j)} ), \dots ).$
          }
        \]
      }

      \State{%
        Let
        \begin{align*}
          \vectgr{\Upsilon}^{(j)}
          & \coloneqq ( \vectgr{\theta}^{(j)},
            \vectgr{\mu}_{1}^{(j)}, \dots,
            \vectgr{\mu}_{K}^{(j)},
            \vectgr{\Gamma}_{1}^{(j)}, \dots,
            \vectgr{\Gamma}_{K}^{(j)} ) \coloneqq -\nabla
            \mathscr{L}(\vectgr{\Omega}^{(j)}) \,.
        \end{align*}
      }

      \State{%
        Find the smallest $M_\textnormal{a} \in \IntegerPP$
        such that
        \begin{align*}
          & \mathscr{L}(\vectgr{\Omega}^{(j)}) -
            \mathscr{L} \left( R_{\vectgr{\Omega}^{(j)}} (
            \bar{\alpha} \beta^{M_\textnormal{a}}
            \vectgr{\Upsilon}^{(j)} ) \right) \\
          & \geq -\sigma \innerp{\nabla
            \mathscr{L} ( \vectgr{\Omega}^{(j)}
            )}{\bar{\alpha} \beta^{M_\textnormal{a}}
            \vectgr{\Upsilon}^{(j)}
            }_{\vectgr{\Omega}^{(j)}} \,.
        \end{align*}
      }

      \State{%
        Define the step-size $t_j^{\textnormal{A}}
        \coloneqq \bar{\alpha} \beta^{M_\textnormal{a}}$.
      }

      \State{%
        Update $\vectgr{\Omega}^{(j+1)} \coloneqq
        R_{\vectgr{\Omega}^{(j)}} (t_j^{\textnormal{A}}
        \vectgr{\Upsilon}^{(j)} )$ via~\eqref{eq:retraction}.
      }

    \EndFor

    \State{%
      $\vectgr{\Omega}_{n+1} \coloneqq
      \vectgr{\Omega}^{(J)}$.
    }

  \end{algorithmic}
  \caption{Solving~\eqref{parameters.task}}
  \label{algo:armijo}
\end{algorithm}

To run the steepest gradient descent in Riemannian
optimization, the \textit{retraction mapping}\/
$R_{ \vectgr{\Omega} }$~\cite{Absil:OptimManifolds:08} is
needed, where, loosely speaking, $R_{ \vectgr{\Omega} }$ is
a mapping which maps an element of the tangent space
$T_{ \vectgr{\Omega} } \mathscr{M}$ to an element in
$\mathscr{M}$. The most celebrated retraction is the
\textit{Riemannian exponential
  mapping}~\cite{Absil:OptimManifolds:08,
  RobbinSalamon:22}. Motivated by this fact, for
$\vectgr{\Omega} \coloneqq ( \vectgr{\xi}, \vect{m}_1,
\dots, \vect{m}_K, \vect{C}_1, \dots, \vect{C}_K ) \in
\mathscr{M}$, for a tangent vector
$\vectgr{\Upsilon} \coloneqq ( \vectgr{\theta},
\vectgr{\mu}_1, \dots, \vectgr{\mu}_K, \vectgr{\Gamma}_1,
\dots, \vectgr{\Gamma}_K) \in T_{ \vectgr{\Omega} }
\mathscr{M}$, and for the step size $t^{\text{A}} > 0$, met
in \cref{algo:armijo}, the retraction mapping
$R_{\vectgr{\Omega}}( t^{\text{A}} \vectgr{\Upsilon} ) = (
R_{\vectgr{\xi}}( t^{\text{A}} \vectgr{\theta} ), \dots,
R_{\vect{m}_k}( t^{\text{A}} \vectgr{\mu}_k ), \dots,
R_{\vect{C}_k}( t^{\text{A}} \vectgr{\Gamma}_k ), \dots )$
is provided by the following:
$\forall k \in \Set{1,\dots, K}$,
\begin{subequations}\label{eq:retraction}
  \begin{align}
    R_{\vectgr{\xi}}( t^{\text{A}} \vectgr{\theta} )
    & \coloneqq
      \vectgr{\xi} + t^{\text{A}} \vectgr{\theta} \,, \\
    R_{\vect{m}_k}( t^{\text{A}} \vectgr{\mu}_k )
    & \coloneqq \vect{m}_k +
      t^{\text{A}} \vectgr{\mu}_k \,, \\
    R_{\vect{C}_k}( t^{\text{A}} \vectgr{\Gamma}_k )
    & \coloneqq \exp_{\vect{C}_k}^{ \text{BW} } (
                   t^{\text{A}} \vectgr{\Gamma}_k ) \,,
  \end{align}
\end{subequations}
where, under the BW metric,
\begin{align*}
  \exp^{\textnormal{BW}}_{ \vect{C}_k } ( t^{\text{A}}
  \vectgr{\Gamma}_k ) \coloneqq \vect{C}_k + t^{\text{A}}
  \vectgr{\Gamma}_k + L_{ \vect{C}_k } ( t^{\text{A}}
  \vectgr{\Gamma}_k )\, \vect{C}_k\,
  L_{ \vect{C}_k } ( t^{\text{A}} \vectgr{\Gamma}_k ) \,.
\end{align*}

\section{Numerical Tests}\label{sec:tests}

\begin{figure}[t]
    \centering
    \subfloat[``Continuous'' loss]{
        \includegraphics[ width = .225\textwidth,
        ]{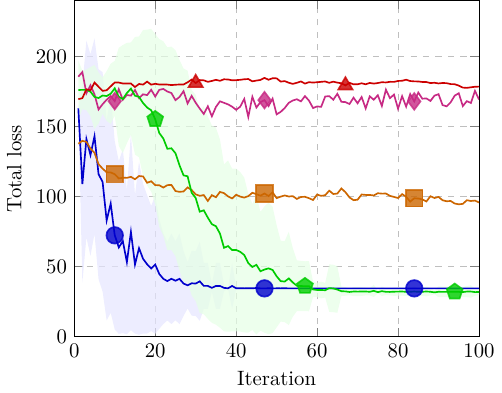}
        \label{fig:pendulum-continuous}
    }
    \subfloat[``Discrete'' loss]{
        \includegraphics[ width = .225\textwidth,
        ]{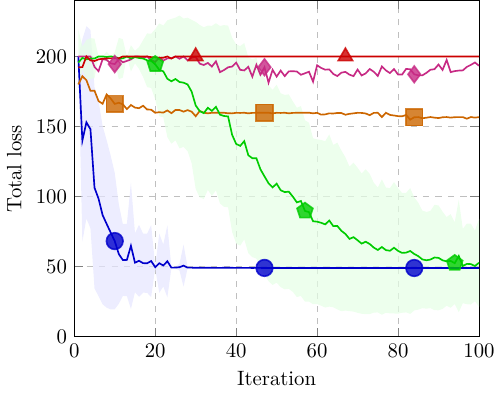}
        \label{fig:pendulum-discrete}
    }
    \caption[]{Inverted-pendulum dataset. Curve markers:
      \cref{algo:PI} with $K=5$:~\tikz{ \node[mark
        size=4pt, blue!80!black, line width = 1pt, opacity
        =.8]{\pgfuseplotmark{*}}; },
      KLSPI~\cite{xu07klspi}:~\tikz{ \node[mark size=4pt,
        orange!80!black, line width = 1pt, opacity
        =.8]{\pgfuseplotmark{square*}}; },
      OBR~\cite{onlineBRloss:16}:~\tikz{ \node[mark
        size=4pt, magenta!80!black, line width=1pt,
        opacity=.8]{\pgfuseplotmark{diamond*}}; },
      DQN~\cite{mnih13dqn}:~\tikz{ \node[mark size=4pt,
        green!80!black, line width=1pt,
        opacity=.8]{\pgfuseplotmark{pentagon*}}; },
      EM-GMMRL~\cite{agostini17gmmrl}:~\tikz{ \node[mark
        size=4pt, red!80!black, line width=1pt,
        opacity=.8]{\pgfuseplotmark{triangle*}}; }.
    }
    \label{fig:pendulum-exp}
\end{figure}

\begin{figure}[t]
    \centering
    \subfloat[``Continuous'' loss]{
        \includegraphics[ width = .225\textwidth,
        ]{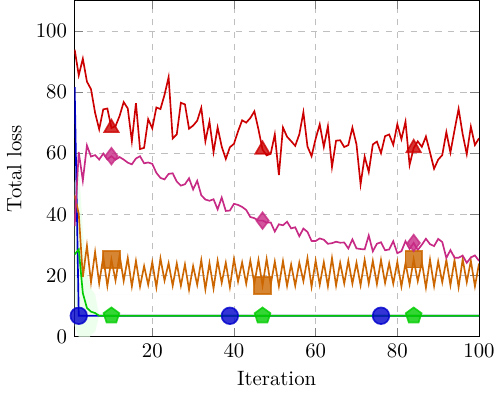}
        \label{fig:mountaincar-continuous}
    }
    \subfloat[``Discrete'' loss]{
        \includegraphics[ width = .225\textwidth,
        ]{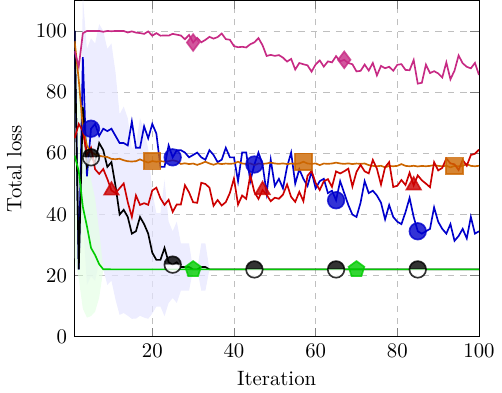}
        \label{fig:mountaincar-discrete}
    }
    \caption[]{Mountain-car dataset. Curve
      markers: \cref{algo:PI} with $K=500$:~\tikz{
        \node[mark size=4pt, black, line width = 1pt,
        opacity =.8]{\pgfuseplotmark{halfcircle*}}; },
      others follow \cref{fig:pendulum-exp}.
}
    \label{fig:mountaincar-exp}
\end{figure}

\begin{figure}
    \centering
    \includegraphics[width = 0.75\columnwidth]
    {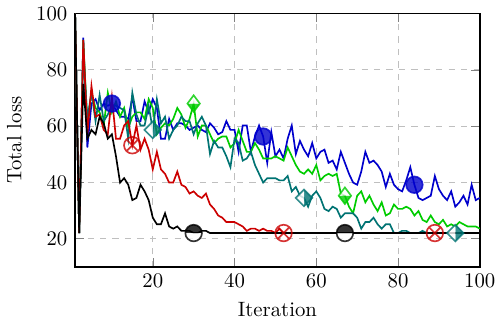}
    \caption[]{%
      Effect of different $K$ in \cref{algo:PI} for
      the setting of~\cref{fig:mountaincar-discrete}. Curve
      markers: $K=20$:~\tikz{ \node[mark size=4pt,
        green!80!black, line width = 1pt, opacity
        =.8]{\pgfuseplotmark{halfdiamond*}}; },
      $K=50$:~\tikz{ \node[mark size=4pt, teal!90!black,
        line width = 1pt, opacity
        =.8]{\pgfuseplotmark{halfsquare right*}}; },
      $K=200$:~\tikz{ \node[mark size=4pt, red, line width =
        1pt, opacity =.8]{\pgfuseplotmark{otimes}}; }. The
      curve markers for $K=5$ and $K=500$ follow those
      of~\Cref{fig:pendulum-exp,fig:mountaincar-exp}. The
      larger the $K$, the richer the hyperparameter space
      $\mathscr{M}$ and the faster the agent learns through
      the feedback from the environment, at the expense of
      increased computational complexity.%
    }
    \label{fig:mountaincar-compare-K}
\end{figure}

Two classical benchmark RL tasks, the \textit{Inverted
  Pendulum}~\cite{doya00rl} and the \textit{Mountain
  Car}~\cite{moore90}, are selected to validate the
proposed~\cref{algo:PI} against:
\begin{enumerate*}[label=\textbf{(\roman*)}]
\item Kernel-based least-squares policy iteration
  (KLSPI)~\cite{xu07klspi}, which utilizes LSTD in RKHS;
\item online Bellman residual (OBR)~\cite{onlineBRloss:16};
\item the popular deep Q-network (DQN)~\cite{mnih13dqn},
  which uses deep neural networks to train the
  $Q$-functions (experienced data are required); and
\item the GMM-based RL~\cite{agostini17gmmrl} via an online
  EM algorithm (EM-GMMRL).
\end{enumerate*}
Two scenarios for the one-step loss function $g$ are also
considered, one where $g$ is continuous and another where it
is discrete. The validation criterion (vertical axes in
\Cref{fig:pendulum-exp,fig:mountaincar-exp,fig:mountaincar-compare-K}) 
measures the
total loss the agent suffers until it achieves the ``goal''
of the task when operating under the current policy $\mu_n$,
with $n$ being the iteration index of~\cref{algo:PI} as well
as the coordinate of the horizontal axes in
\Cref{fig:pendulum-exp,fig:mountaincar-exp,fig:mountaincar-compare-K}. 
Results are
averages from \num{100} independent tests. Software code was
written in Julia~\cite{julia17}/Python.

The ``inverted pendulum''~\cite{doya00rl} refers to the
problem of swinging up a pendulum from its lowest position
to the upright one, given a limited number of torques.
The state
$\vect{s} \coloneqq [\theta, \dot{\theta}]^{\intercal}$,
where $\theta\in [-\pi, \pi]$ is the angular position
($\theta = 0$ corresponds to the upright position), and
$\dot{\theta}\in [-4, 4]\text{s}^{-1}$ is the angular
velocity. The action space is the set of torques
$\mathfrak{A} \coloneqq \Set{-5, -3, 0, 3, 5}\text{N}$. The
continuous one-step loss is defined as
$g(\vect{s},a) \coloneqq |\theta|/\pi$, while the discrete
one is defined as $g(\vect{s}, a) \coloneqq 0$, if
$\theta = 0$, and $g(\vect{s}, a) \coloneqq 1$, if
$\theta \neq 0$.

To collect data samples $\mathscr{D}_{\mu_n}$
in~\cref{algo:PI}, the pendulum starts from an angular
position and explores a number of actions under the current
policy $\mu_n$. This exploration is called an episode, and
per iteration $n$ in ~\cref{algo:PI}, data
$\mathscr{D}_{\mu_n}$ with
$T \coloneqq ( \text{number of episodes}) \times
(\text{number of actions}) = 20 \times 70 = 1400$ are
collected.
KLSPI~\cite{xu07klspi} and OBR~\cite{onlineBRloss:16} use
the Gaussian kernel with bandwidth $\sigma_\kappa = 2$,
while their ALD threshold is
$\delta_{\textnormal{ALD}}=0.01$. KLSPI and OBR need
$T = 5000$ to reach their ``optimal'' performance for the
task at hand. DQN~\cite{mnih13dqn} uses a fully-connected
neural network with \num{2} hidden layers of size \num{128},
with batch size of \num{64}, and a replay buffer (experienced
data) of size \num{1e5}. For
EM-GMMRL~\cite{agostini17gmmrl}, $T = 500$, while its
threshold to add new Gaussian functions in its dictionary is
$10^{-4}$. 

It can be seen from~\cref{fig:pendulum-exp},
that the proposed~\cref{algo:PI} scores the best performance
with no use of replay buffer (experienced data), unlike
DQN~\cite{mnih13dqn} which requires a large replay buffer,
and exhibits slower learning speed and higher variance than
GMM-QFs. KLSPI~\cite{xu07klspi} underperforms, while
OBR~\cite{onlineBRloss:16} and
EM-GMMRL~\cite{agostini17gmmrl} fail to score a satisfactory
performance. Notice that KLSPI and OBR are given more
exploration data than~\cref{algo:PI}. It is also worth
noting here that EM algorithms are sensitive to
initialization~\cite{Figueiredo:mixtures:02}, and that
several initialization strategies were tried in all of the
numerical tests.

``Mountain car''~\cite{moore90} refers to the task of
accelerating a car to reach the top of the hill from the
bottom of a sinusoidal valley, where the slope equation is
given by $y = \sin(3x)$ in the $xy$-plane, with
$x\in [-1.2, 0.6]$. The state
$\vect{s}\coloneqq [x, v]^{\intercal}$, where the velocity
of the car $v\in [-0.07, 0.07]$. The goal is achieved when
the car gets beyond $x_g \coloneqq 0.5$ with velocity larger
than or equal to $v_g \coloneqq 0$, that is, whenever the
car reaches a state in
$\mathfrak{S}_g \coloneqq \Set{ [x, v]^{\intercal} \given x
  \geq x_g, v \geq v_g }$. The discrete one-step loss is
defined as $g(\vect{s}, a) \coloneqq 1$, if
$\vect{s} \notin \mathfrak{S}_g$, while
$g(\vect{s}, a) \coloneqq 0$, if
$\vect{s} \in \mathfrak{S}_g$. The continuous one-step loss
is defined as
$g(\vect{s}, a) \coloneqq [ \max (x_g - x, 0) +
\max(v_g - v, 0) ] / 2$.

With regards to the data $\mathcal{D}_{\mu_n}$
in~\cref{algo:PI}, a strategy similar to that of the
inverted pendulum is used. More specifically, $T = 1000$ for
the proposed GMM-QFs, while $T = 20000$ for
KLSPI~\cite{xu07klspi} and $T = 1000$ for
OBR~\cite{onlineBRloss:16}. A Gaussian kernel with width of
$\sigma_\kappa = 0.1$ is used for KLSPI~\cite{xu07klspi} and
OBR~\cite{onlineBRloss:16}. The implementation of
DQN~\cite{mnih13dqn} is identical to one for the
inverted-pendulum case, while $T = 100$ for
EM-GMMRL~\cite{agostini17gmmrl}.

The proposed GMM-QFs outperform all competing methods
in~\cref{fig:mountaincar-continuous}, while
DQN~\cite{mnih13dqn} scores the best performance in
\cref{fig:mountaincar-discrete}. Note again here that DQN
uses a large number of experienced data (size of replay buffer
is \num{1e5}), while the proposed GMM-QFs needs \text{no}\/
experienced data to achieve the performance of
\cref{fig:mountaincar-exp}. Observe also that by increasing
the number $K$ of Gaussians in~\eqref{eq:gmm-q}, GMM-QFs
reach the total-loss performance of DQN in
\cref{fig:mountaincar-discrete}, at the expense of increased
computational complexity; see also
\cref{fig:mountaincar-compare-K}.
OBR~\cite{onlineBRloss:16} and
EM-GMMRL~\cite{agostini10gmmrl} perform better here than in
\cref{fig:pendulum-exp}, with the EM-GMMRL agent showing
better ``learning abilities'' than the OBR one in
\cref{fig:mountaincar-discrete}, but vice versa in
\cref{fig:mountaincar-continuous}.

\section{Conclusions}\label{sec:conslusions}

This paper established the novel class of GMM Q-functions
(GMM-QFs), and offered a Riemannian-optimization algorithm,
to learn the hyperparameters of GMM-QFs, as a novel
policy-evaluation step in a policy-iteration scheme for
computing optimal policies. The proposed design shows ample
degrees of freedom not only because it introduces a rich
hyperparameter space, but also because it establishes the
exciting connection between Q-function identification in RL
and the powerful toolbox of Riemannian
optimization. Numerical tests on benchmark tasks
demonstrated the superior performance of the proposed design
over state-of-the-art schemes.

\clearpage
\printbibliography[title = {\normalsize\uppercase{References}}]
\end{document}

%% file: MinhPreamble.tex
\usepackage{spconf}

\usepackage[T1]{fontenc}

\usepackage{calrsfs}
\usepackage{amsmath, amsthm, accents, amssymb, amsfonts, bm,
  mathtools, cases}
\usepackage[cal = boondox,%
calscaled = 1.1,%
bb = ams,%
bbscaled = .94,%
frakscaled = .97,%
scr = esstix]{mathalpha}

\usepackage{algorithm}
\usepackage{algpseudocode}
\usepackage[inline]{enumitem}
\usepackage{graphicx}
\usepackage{hyperref}
\usepackage{booktabs}

\usepackage{subcaption}

\usepackage{textcomp}
\usepackage{xcolor}
\usepackage{etoolbox}
\def\BibTeX{{\rm B\kern-.05em{\sc i\kern-.025em b}\kern-.08em
    T\kern-.1667em\lower.7ex\hbox{E}\kern-.125emX}}
\usepackage[capitalize]{cleveref}
\usepackage{url}

\usepackage{siunitx}
\usepackage[textsize=footnotesize]{todonotes}

\usepackage{pgfplots}
\pgfplotsset{compat=newest}
\usepgflibrary[plotmarks]
\usepgfplotslibrary{fillbetween}
\usetikzlibrary{angles, arrows, arrows.meta, shapes, tikzmark, patterns}

\newtheorem{proposition}{Proposition}

\newcommand\given{{\mathbin{}\mid\mathbin{}}}
\newcommand{\IntegerP}{\mathbb{N}}
\newcommand{\IntegerPP}{\mathbb{N}_*}
\newcommand{\Real}{\mathbb{R}}

\newcommand{\PS}{\mathbb{S}_{++}}
\newcommand\SetSymbol[1][]{
  \nonscript\,#1\vert \allowbreak \nonscript\,\mathopen{}}
\DeclarePairedDelimiterX\Set[1]{\lbrace}{\rbrace}%
{ \renewcommand\given{\SetSymbol[\delimsize]} #1 }
\DeclarePairedDelimiterX\norm[1]\lVert\rVert{\ifblank{#1}{\:\cdot\:}{#1}}
\DeclarePairedDelimiterX\innerp[2]{\langle}{\rangle}{#1
  \mathop{}\delimsize\vert\mathop{} #2}

\newcommand\vect[1]{\mathbf{#1}}
\newcommand\vectgr[1]{\bm{#1}}

\newcommand{\Banach}{\mathcal{B}}

\DeclareMathOperator{\Fix}{Fix}

\DeclareMathOperator{\trace}{tr}

\crefname{theorem}{Theorem}{Theorems}
\crefname{lemma}{Lemma}{Lemmas}
\crefname{thmlisti}{Theorem}{Theorems}
\crefname{algo}{Algorithm}{Algorithms}
\crefname{figure}{Figure}{Figures}
\crefname{section}{Section}{Sections}

\newlist{thmlist}{enumerate}{1}
\setlist[thmlist]{label=\textbf{(\roman{*})}, ref=\thetheorem(\roman{*}), noitemsep}

\hypersetup{hidelinks}
\allowdisplaybreaks

\usepackage[%
backend = biber,%
maxbibnames = 10,%
minbibnames = 2,%
style = ieee,%
citestyle = numeric-comp,%
bibencoding = utf8,%
datamodel = software,%
defernumbers = true,%
sortcites=true,%
doi=true,%
isbn=false,%
url=false,%
eprint=true%
]{biblatex}

